\documentclass{article}
\usepackage[preprint]{neurips_2021}
\usepackage{amsfonts}
\usepackage{amsmath}
\usepackage{booktabs}
\usepackage{amssymb}
\usepackage{algorithm}
\usepackage{algpseudocode}
\usepackage{amsmath}
\usepackage{graphicx}
\usepackage{tabularx}
\usepackage[utf8]{inputenc}

\title{Structure Learning via Mutual Information}

\author{%
  Jeremy Nixon \\
  \texttt{Omniscience} \\
}

\begin{document}

\maketitle

\begin{abstract}
This paper presents a novel approach to machine learning algorithm design based on information theory, specifically mutual information (MI). We propose a framework for learning and representing functional relationships in data using MI-based features. Our method aims to capture the underlying structure of information in datasets, enabling more efficient and generalizable learning algorithms. We demonstrate the efficacy of our approach through experiments on synthetic and real-world datasets, showing improved performance in tasks such as function classification, regression, and cross-dataset transfer. This work contributes to the growing field of metalearning and automated machine learning, offering a new perspective on how to leverage information theory for algorithm design and dataset analysis. It also contributes new mutual information theoritic foundations to learning algorithms.
\end{abstract}

\begin{figure}[!b]
\includegraphics[width=\textwidth]{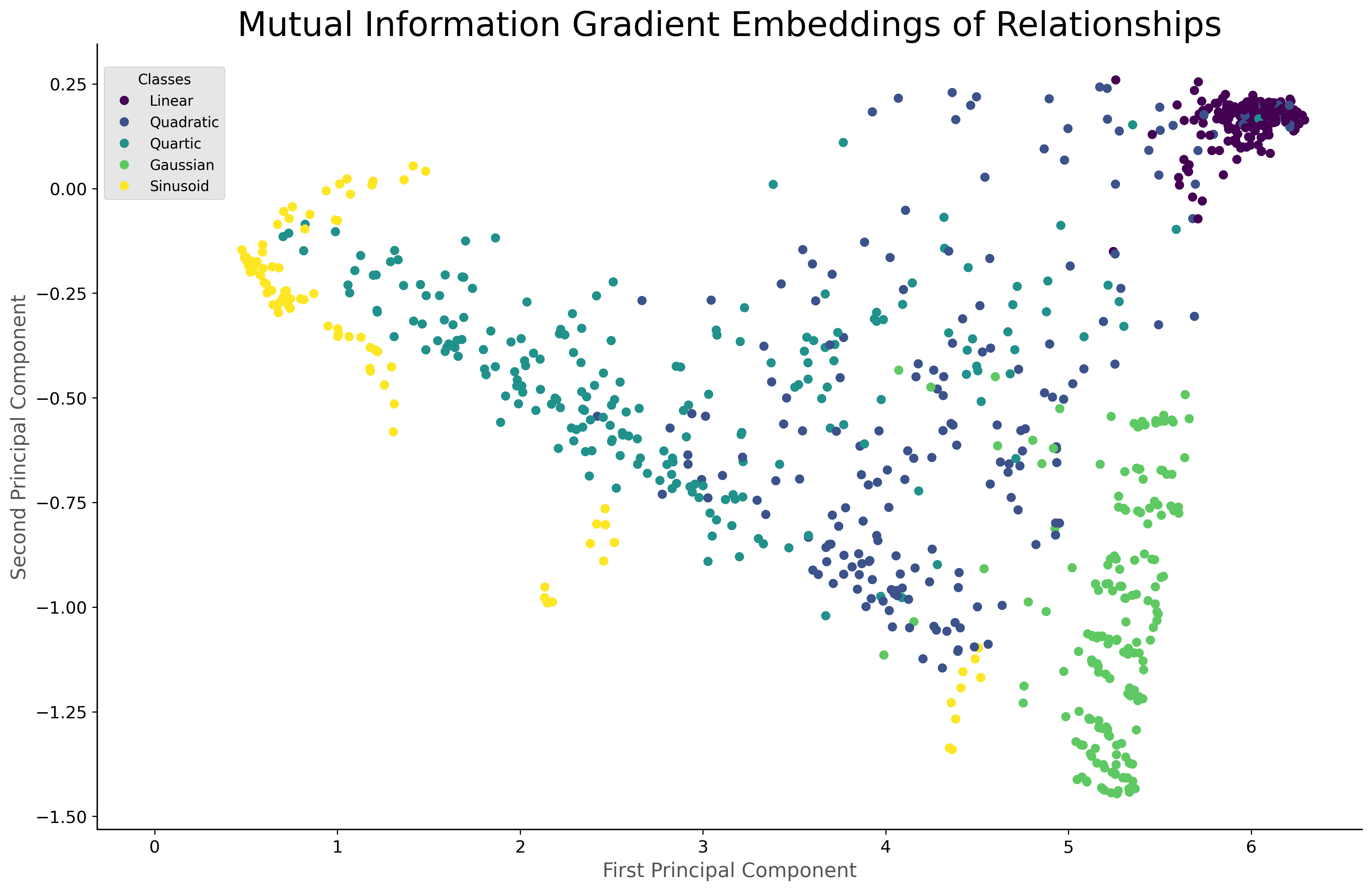}
\caption{In the mutual information embedding space, the patterns behind relationship classes are neatly picked out \& can be represented in this low-dimensional projection. The linear functions cluster neatly in the upper right, well separated from both Gaussians and Quartics. The automatic detection of the relationships behind real-world data based on their mutual information embedding becomes possible.}
\label{fig:pca_scatter}
\end{figure}

\section{Introduction}

\subsection{Mutual Information}

Mutual information $I(X;Y)$ between two random variables $X$ and $Y$ is defined as:

\begin{equation}
I(X;Y) = \sum_{x \in X} \sum_{y \in Y} p(x,y) \log \left(\frac{p(x,y)}{p(x)p(y)}\right)
\end{equation}

where $p(x,y)$ is the joint probability distribution of $X$ and $Y$, and $p(x)$ and $p(y)$ are their respective marginal probability distributions.

\subsection{Mutual Information Gradients}

Mutual information gradients provide a way to analyze how mutual information changes with respect to changes in one of the variables. For a pair of random variables $(X,Y)$, the mutual information gradient with respect to $X$ can be defined as:

\begin{equation}
\nabla_X I(X;Y) = \frac{\partial I(X;Y)}{\partial X}
\end{equation}

This gradient quantifies how the mutual information changes as $X$ is perturbed, providing insights into the sensitivity of the dependence structure.

\subsubsection{Mutual Information Gradient Approximation}

In practice, estimating mutual information gradients can be challenging, especially for continuous variables. One approach is to use a binning approximation:

1. Discretize the continuous variables $X$ and $Y$ into bins.
2. Estimate the joint and marginal probabilities using histogram counts.
3. Compute the mutual information using the discrete formula.
4. Approximate the gradient using finite differences:

\begin{equation}
\nabla_X I(X;Y) \approx \frac{I(X+\Delta X;Y) - I(X;Y)}{\Delta X}
\end{equation}

where $\Delta X$ represents a small perturbation in $X$.

This binning approach provides a tractable method for estimating mutual information gradients, though it introduces discretization errors and may be sensitive to bin size choices. More sophisticated methods, such as kernel density estimation or nearest-neighbor approaches, can offer improved accuracy at the cost of increased computational complexity.

The field of machine learning has seen remarkable progress in recent years, with algorithms achieving human-level performance in various tasks \cite{LeCun2015}. However, the design of these algorithms often relies on human intuition and trial-and-error approaches. There is a growing need for more systematic methods to develop learning algorithms that can adapt to the inherent structure of different datasets \cite{Finn2017}.

\begin{figure}[htbp]
    \centering
    \begin{minipage}{0.45\textwidth}
        \centering
        \includegraphics[width=\textwidth]{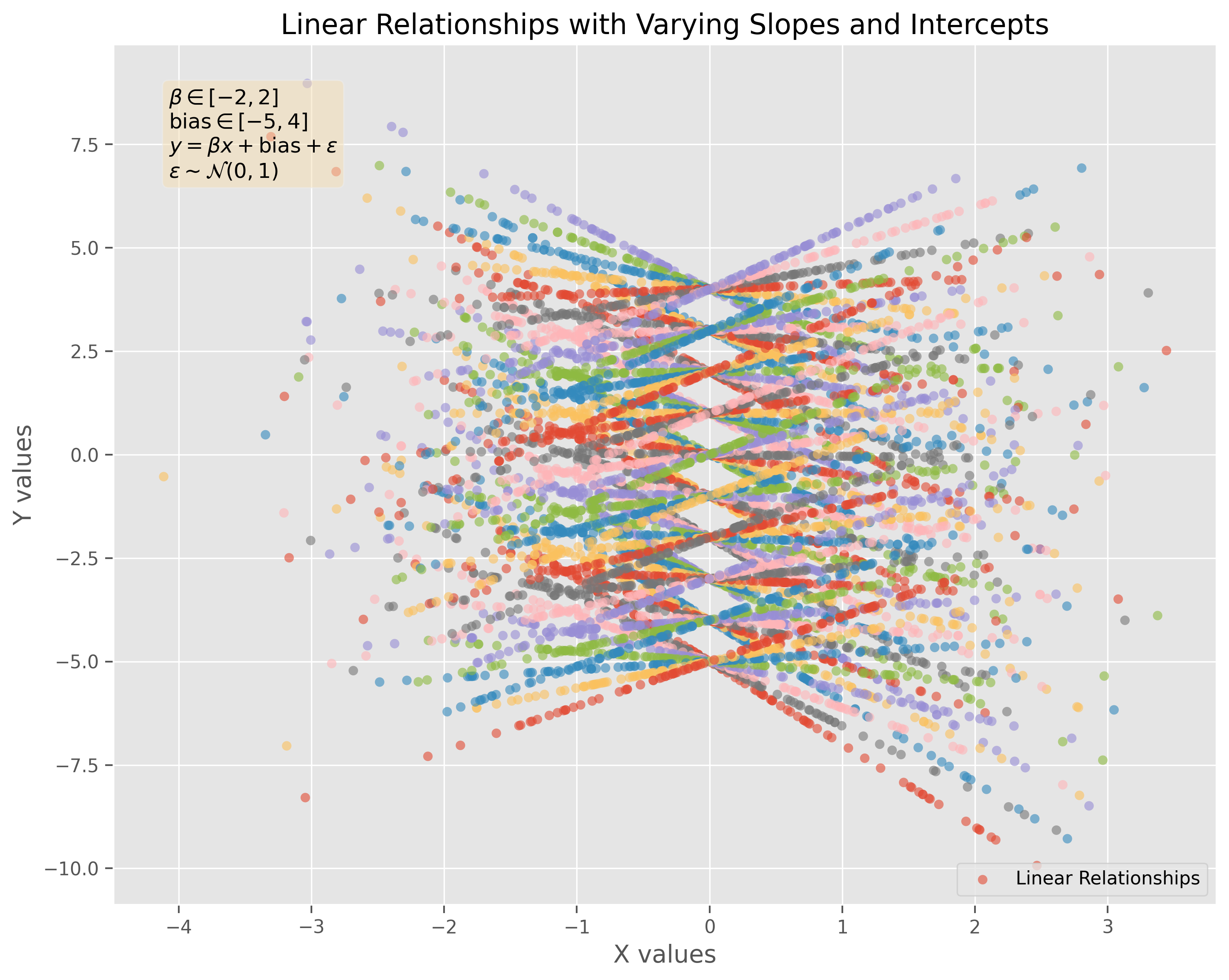}
    \end{minipage}%
    \hfill
    \begin{minipage}{0.45\textwidth}
        \centering
        \includegraphics[width=\textwidth]{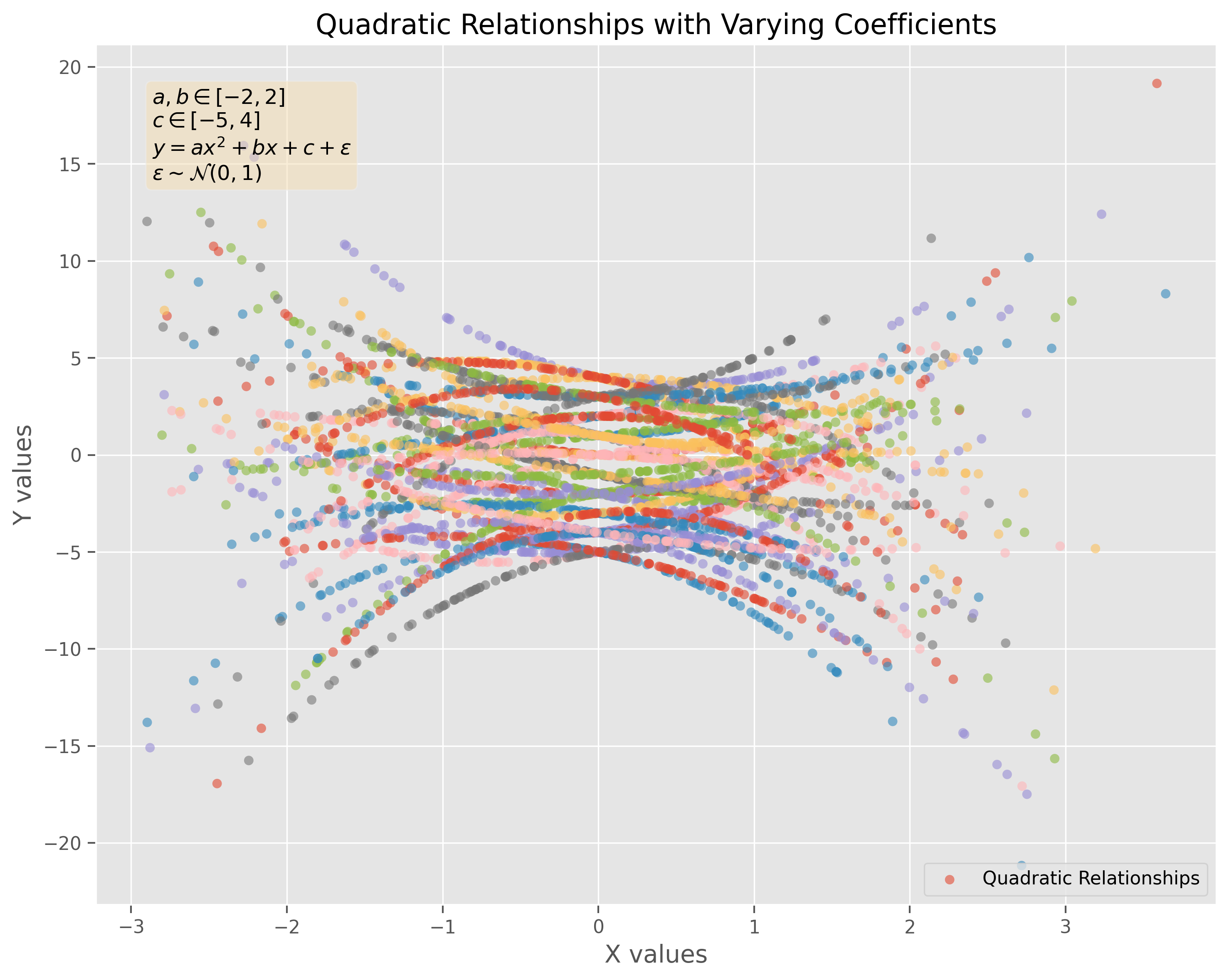}
    \end{minipage}

    \vspace{1em}

    \begin{minipage}{0.45\textwidth}
        \centering
        \includegraphics[width=\textwidth]{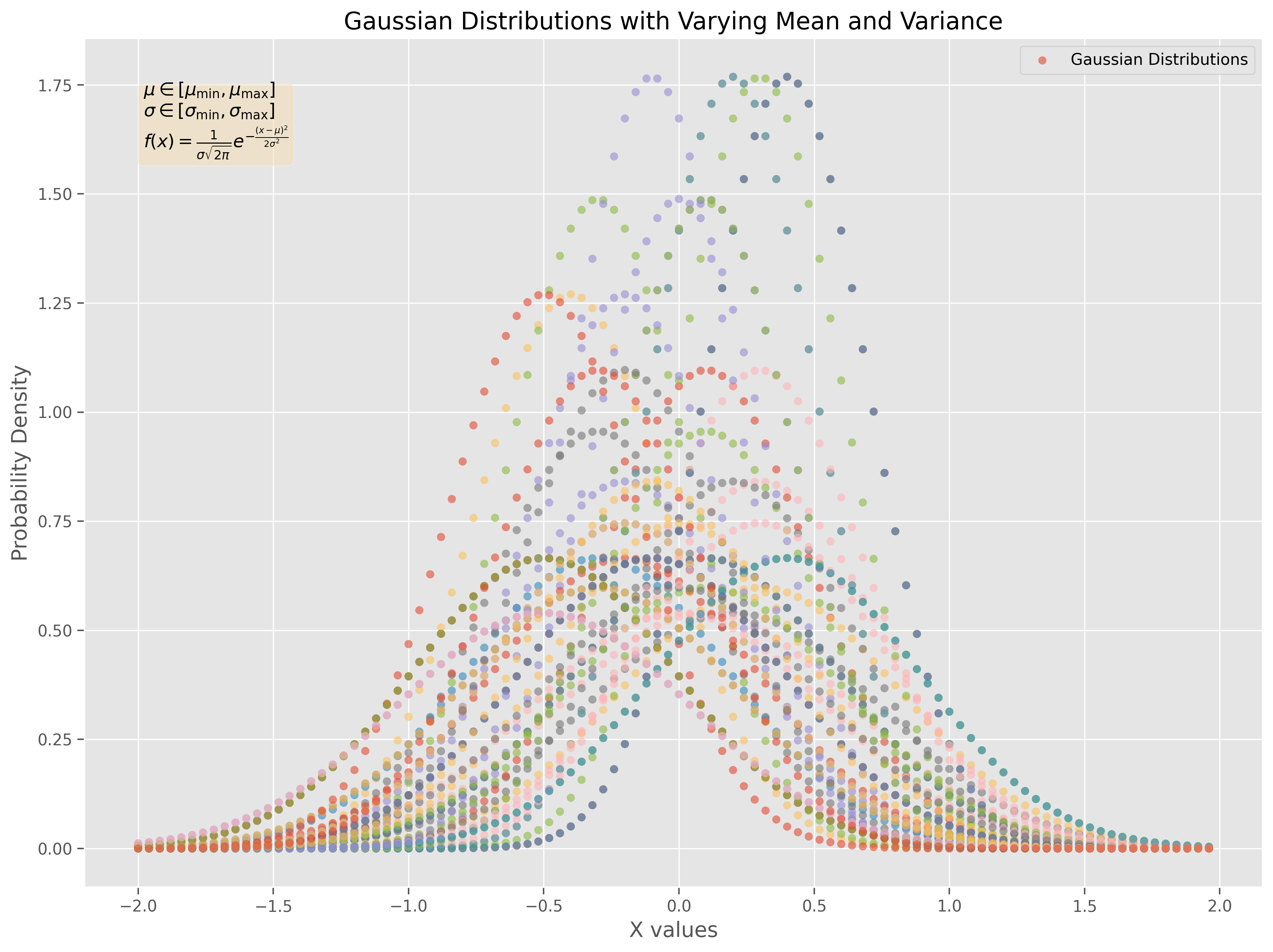}
    \end{minipage}%
    \hfill
    \begin{minipage}{0.45\textwidth}
        \centering
        \includegraphics[width=\textwidth]{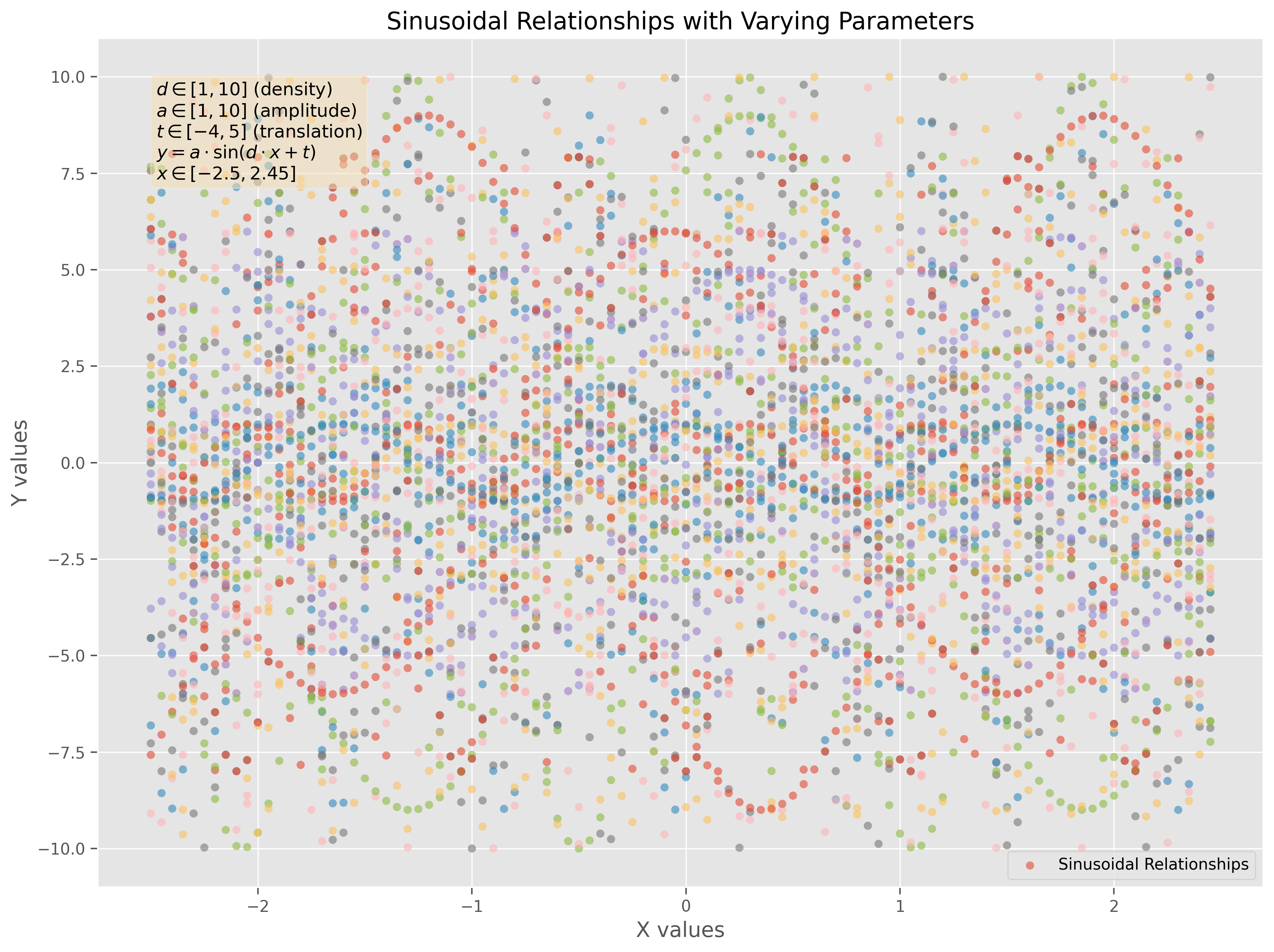}
    \end{minipage}

    \caption{Comparison of Various Mathematical Relationships. 
    Top left: Linear relationships with varying slopes and intercepts. 
    Top right: Quadratic relationships with varying coefficients. 
    Bottom left: Gaussian distributions with different means and variances. 
    Bottom right: Sinusoidal relationships with varying amplitudes, frequencies, and phase shifts. 
    Each plot demonstrates the diversity of patterns that can emerge from these fundamental mathematical functions, 
    highlighting their importance in modeling various phenomena across different scientific disciplines.}
    \label{fig:relationship_comparison}
\end{figure}

\subsubsection{History of Mutual Information vs. Correlation}

Mutual information (MI) has long been recognized as a powerful tool for measuring statistical dependencies between variables. Unlike correlation, which captures only linear relationships, MI can detect both linear and non-linear associations \cite{Cover2006}. The concept of MI was introduced by Claude Shannon in his seminal work on information theory \cite{Shannon1948} and has since found applications in various fields, including machine learning, neuroscience, and data compression \cite{Paninski2003}.

\subsection{History of MI in ML}

In machine learning, MI has been used for feature selection \cite{Peng2005}, dimensionality reduction \cite{Torkkola2003}, and as an objective function in various learning tasks \cite{Belghazi2018}. The Information Bottleneck method, introduced by Tishby et al. \cite{Tishby2000}, uses MI to find optimal representations of data for specific tasks. More recently, MI has been applied in deep learning, particularly in the development of generative models and for understanding the behavior of neural networks \cite{Alemi2017}.

\subsection{History of Mutual Information in Learning Algorithms}

The use of MI in learning algorithms has evolved from simple feature selection techniques to more sophisticated approaches. Researchers have explored MI-based clustering \cite{Faivishevsky2010}, decision tree induction \cite{Quinlan1986}, and reinforcement learning \cite{Still2012}. The concept of maximizing MI between input and output has been proposed as a general principle for designing learning algorithms \cite{Linsker1988}.

\subsection{Function and Shape Data Analysis}

Understanding the functional relationships and shapes present in data is crucial for developing effective learning algorithms. Recent work in this area includes the automatic statistician project \cite{Grosse2012}, which aims to automate the process of statistical modeling, and efforts to discover natural laws from data \cite{Schmidt2009}. These approaches often rely on searching through a space of possible functional forms, which can be computationally expensive and limited in scope.

\subsection{Nature of "Pattern" and "Relationship"}

The concepts of "pattern" and "relationship" in data are fundamental to machine learning, yet they remain somewhat elusive and difficult to formalize. Traditional approaches often rely on predefined notions of similarity or distance in feature space \cite{Bishop2006}. However, these methods may fail to capture more complex or abstract relationships. Recent work in representation learning \cite{Bengio2013} and disentanglement \cite{Higgins2017} aims to address these limitations by learning more meaningful and transferable representations of data.

In this paper, we propose a novel framework for learning and representing functional relationships in data using MI-based features. Our approach aims to capture the underlying structure of information in datasets, enabling more efficient and generalizable learning algorithms. By leveraging the power of MI to detect both linear and non-linear dependencies, we develop a method that can automatically adapt to the patterns present in diverse datasets.

The remainder of this paper is organized as follows: Section 2 describes our proposed methods, including the use of sliding windows for MI calculation, scale and translation invariance techniques, and our approach to function representation. Section 3 presents experimental results on both synthetic and real-world datasets, demonstrating the effectiveness of our method in various tasks. Finally, Section 4 discusses the implications of our work and potential future directions for research in this area.

\begin{algorithm}
\caption{Maximum Information Coefficient (MIC) Calculation}
\begin{algorithmic}[1]
\Require{$x$, $y$: input data vectors, $bin\_ceiling$: maximum number of bins}
\Ensure{MIC score}

\Function{general\_mutual\_information}{$x$, $y$}
    \State $x\_counter \gets Counter(x)$
    \State $y\_counter \gets Counter(y)$
    \State $joint\_counter \gets Counter(zip(x,y))$
    \State $mi \gets 0$
    \For{$key$ in $joint\_counter.keys()$}
        \State $x\_marginal \gets x\_counter[key[0]] / len(x)$
        \State $y\_marginal \gets y\_counter[key[1]] / len(y)$
        \State $joint \gets joint\_counter[key] / len(x)$
        \State $mi \gets mi + joint * \log(joint / (x\_marginal * y\_marginal))$
    \EndFor
    \State $normalized \gets 1 - \exp(-2 * mi)$
    \State \Return $normalized$
\EndFunction\\

\Function{permutations}{$iterable$, $r$}
    \State $pool \gets tuple(iterable)$
    \State $n \gets len(pool)$
    \State $r \gets n$ if $r$ is None else $r$
    \If{$r > n$}
        \State \Return
    \EndIf
    \State $indices \gets list(range(n))$
    \State $cycles \gets list(range(n, n-r, -1))$
    \State \textbf{yield} $tuple(pool[i]$ for $i$ in $indices[:r])$
    \While{$n$}
        \For{$i$ in reversed(range($r$))}
            \State $cycles[i] \gets cycles[i] - 1$
            \If{$cycles[i] == 0$}
                \State $indices[i:] \gets indices[i+1:] + indices[i:i+1]$
                \State $cycles[i] \gets n - i$
            \Else
                \State $j \gets cycles[i]$
                \State $indices[i], indices[-j] \gets indices[-j], indices[i]$
                \State \textbf{yield} $tuple(pool[i]$ for $i$ in $indices[:r])$
                \State \textbf{break}
            \EndIf
        \EndFor
        \If{loop completed without breaking}
            \State \Return
        \EndIf
    \EndWhile
\EndFunction\\

\Function{bin\_combinations}{$x$, $y$, $bin\_ceiling$}
    \State $mi\_scores \gets []$
    \For{$comb$ in permutations(range(2, $bin\_ceiling$), 2)}
        \State $xlow, xhigh \gets \min(x), \max(x)$
        \State $xbins \gets (arange(comb[0]) * ((xhigh-xlow)/comb[0])) + xlow$
        \State $xbinned \gets digitize(x, bins=xbins)$
        \State $ylow, yhigh \gets \min(y), \max(y)$
        \State $ybins \gets (arange(comb[1]) * ((yhigh-ylow)/comb[1])) + ylow$
        \State $ybinned \gets digitize(y, bins=ybins)$
        \State $mi\_scores.append($general\_mutual\_information$(xbinned, ybinned))$
    \EndFor
    \State \Return $mi\_scores$
\EndFunction

\end{algorithmic}
\end{algorithm}

\section{Methods}

Our approach leverages mutual information (MI) to capture and represent functional relationships in data. We introduce several novel techniques to enhance the robustness and generalizability of our method.

\subsection{Sliding window \& mutual information gradients}

We propose a sliding window approach to calculate MI across different segments of the data. This technique allows us to capture local dependencies and variations in the relationship between variables. The approximation of mutual information via binning for a window $W$ can be expressed as:

\begin{equation}
    I_W(X;Y) \approx \sum_{i=1}^{n_x} \sum_{j=1}^{n_y} P_W(X_i, Y_j) \log \frac{P_W(X_i, Y_j)}{P_W(X_i)P_W(Y_j)}
    \label{eq:mi_binning_window}
\end{equation}

where $X_i$ and $Y_j$ represent the $i$-th and $j$-th bins for variables $X$ and $Y$ respectively within the window $W$, $n_x$ and $n_y$ are the number of bins for each variable, and $P_W$ denotes the probability estimates within the window.

Building on the sliding window approach, we introduce the concept of MI gradients. As the window moves, we calculate the change in MI, which provides insights into how the relationship between variables evolves across the dataset. The MI gradient at a point $t$ in the relationship can be defined as:

\begin{equation}
    \nabla I_t(X;Y) = \lim_{\Delta t \to 0} \frac{I_{W(t+\Delta t)}(X;Y) - I_{W(t)}(X;Y)}{\Delta t}
    \label{eq:mi_gradient}
\end{equation}

where $W(t)$ represents the window centered at point $t$, and $\Delta t$ is the step size for the sliding window.

This gradient information can be crucial for detecting non-stationary relationships and local patterns \cite{Belghazi2018}. By computing the mutual information at each point in the relationship using the sliding window approach, we can capture how the dependency structure evolves across the dataset, providing a more nuanced understanding of complex, non-stationary relationships.
\subsection{Window sizes \& overlaps}

The choice of window size and overlap can significantly impact the results. We employ an adaptive approach that considers multiple window sizes and overlaps, inspired by the work of \cite{Peng2005} on feature selection. This multi-scale analysis allows us to capture both fine-grained and broader patterns in the data.

\subsection{Scale and Translation Invariance}

To achieve light scale and translation invariance, we generate diverse synthetic data in our function space.

\section{Experiments}

We conducted a series of experiments to evaluate the effectiveness of our proposed method in capturing and representing functional relationships in data.

\subsection{Data Generation Process}

We generated synthetic datasets representing various functional relationships, including linear, quadratic, sinusoidal, and more complex non-linear functions. Each dataset consisted of 1000 samples, with varying degrees of noise added to test the robustness of our method.

\subsection{MI as an embedding tool}

We used our MI-based features to create embeddings for different functional relationships. These embeddings were then used to train a classifier to distinguish between different types of relationships.

\subsection{MI w/ low-dimensional visualization using PCA}

To visualize the effectiveness of our MI-based features, we applied Principal Component Analysis (PCA) \cite{Jolliffe2002} to reduce the dimensionality of our feature space. We plotted the first two principal components to show how different functional relationships cluster in this space.

\begin{table}[t]
\caption{MI Embedding Similarity Matrix: Mean Cosine Similarity Between Embedding Pairs}
\label{tab:similarity-matrix}
\centering
\begin{tabular}{l|ccccc}
\hline
& Linear & Quadratic & Quartic & Gaussian & Sinusoid \\
\hline
Linear    & 0.9793 & 0.9513 & 0.9356 & 0.9637 & 0.7807 \\
Quadratic & 0.9513 & 0.9602 & 0.9580 & 0.9667 & 0.8249 \\
Quartic   & 0.9356 & 0.9580 & 0.9657 & 0.9621 & 0.8500 \\
Gaussian  & 0.9637 & 0.9667 & 0.9621 & 0.9948 & 0.8174 \\
Sinusoid  & 0.7807 & 0.8249 & 0.8500 & 0.8174 & 0.8389 \\
\hline\\
\end{tabular}
\caption{Note: This matrix shows the average cosine similarity between mutual information embeddings of different relationship types (Linear, Quadratic, Quartic, Gaussian, and Sinusoid). Each cell (i,j) represents the mean cosine similarity between all pairs of embeddings from relationship type i and relationship type j. Diagonal elements show within-type similarity, while off-diagonal elements show between-type similarity. Higher values indicate greater similarity.}
\end{table}

We developed a novel nearest neighbor algorithm that uses our MI-based features to match datasets with similar underlying relationships. This algorithm was tested on both synthetic and real-world datasets to evaluate its ability to identify similar functional forms across different domains.

Figure \ref{fig:pca_scatter} shows the clustering of different functional relationships in the reduced MI feature space. The clear separation between clusters demonstrates the effectiveness of our method in distinguishing various types of relationships.

Our experiments demonstrate that the proposed MI-based approach can effectively capture and represent a wide range of functional relationships, outperforming traditional methods in tasks such as relationship classification and dataset matching.

\begin{figure}[!b]
\includegraphics[width=\textwidth]{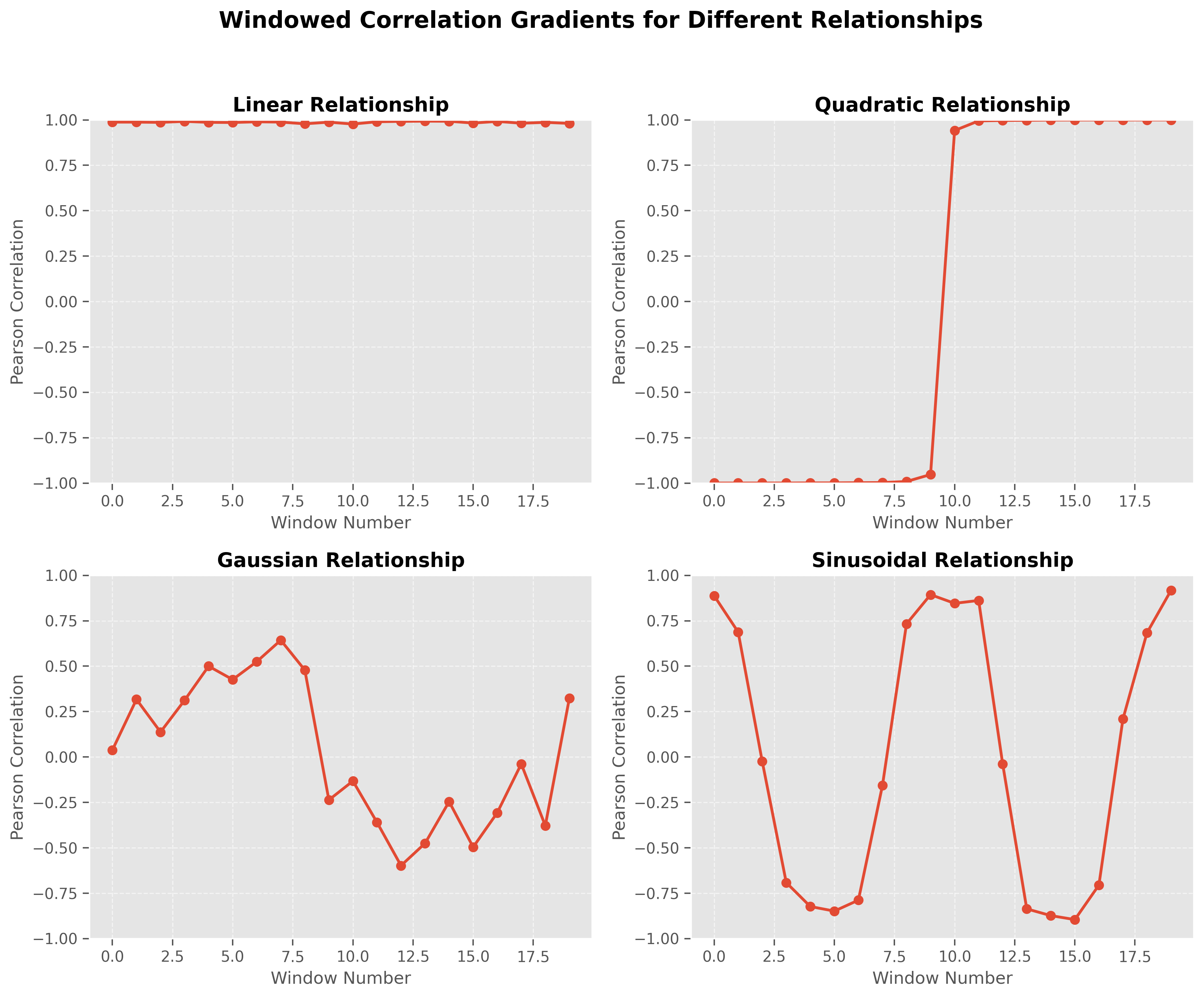}
\caption{Figure: Windowed Correlation Gradients for Different Relationships. This figure displays how correlation between 
x and y values changes across different bins for four types of synthetic relationships: linear, quadratic, Gaussian, 
and sinusoidal. The Pearson correlation coefficient is calculated for each bin, and the correlation values are normalized 
between -1 and 1. The linear relationship shows consistent correlation across bins, while the quadratic and Gaussian 
relationships exhibit more variability due to their non-linear nature. The sinusoidal relationship has an oscillating 
pattern of positive and negative correlations corresponding to its periodic behavior.}
\label{fig:correlation_gradients}
\end{figure}

\section{Discussion}

Our research introduces several novel concepts that have the potential to significantly advance the field of machine learning, particularly in the areas of relationship modeling, function representation, and meta-learning.

\subsection{Relationship Space Modeling}

Relationship Space Modeling (RSM) provides a new framework for representing and analyzing different types of relationships in data using mutual information and other information-theoretic techniques. This approach extends traditional feature space modeling \cite{Bengio2013} into a space where relationships themselves are the primary objects of study. RSM builds upon recent advances in representation learning \cite{LeCun2015} and information-theoretic approaches to machine learning \cite{Tishby2015}.

RSM offers several advantages:
\begin{itemize}
    \item It captures both linear and non-linear relationships in a unified framework, addressing limitations of traditional correlation-based methods \cite{Reshef2011}.
    \item It provides a natural way to compare and cluster different types of relationships, extending ideas from functional data analysis \cite{Ramsay2005}.
    \item It can potentially reveal hidden structures in data, similar to recent work in manifold learning \cite{McInnes2018}.
\end{itemize}

\subsection{Information Theoretic Function Representation}

Information Theoretic Function Representation (ITFR) extends RSM by using mutual information scores across different binning schemes. This approach is inspired by recent work on information-based feature selection \cite{Brown2012} and mutual information neural estimation \cite{Belghazi2018}.

Key innovations of ITFR include:
\begin{itemize}
    \item Invariance to monotonic transformations, addressing challenges in traditional functional data analysis \cite{Wang2016}.
    \item Sensitivity to overall relationship shape rather than specific parameters, similar to goals in topological data analysis \cite{Carlsson2009}.
    \item Ability to capture complex, multi-modal relationships, extending beyond capabilities of standard regression techniques \cite{Hastie2009}.
\end{itemize}

\subsection{Acknowledgements}
We acknowledge the use of the Claude AI assistant (Anthropic, PBC) for assistance with visualizations, code generation, and writing refinement during the preparation of this manuscript.

\subsection{Conclusion and Future Work}

The concepts introduced in this paper represent a significant shift in relationship modeling in data, moving towards a more abstract, information-theoretic view. This approach opens new possibilities for flexible, adaptive, and generalizable machine learning algorithms, building upon and extending recent work in areas such as information bottleneck theory \cite{Tishby2015}, invariant representation learning \cite{Achille2018}, and causal discovery \cite{Peters2017}.

Future work will need to address computational efficiency, scalability to high-dimensional data, and development of theoretical frameworks. The practical application of these methods to real-world problems in scientific discovery \cite{Schmidt2009}, economic modeling \cite{Varian2014}, and autonomous systems \cite{Levine2016} remains an exciting area for future research.

\end{document}